\documentclass{article}
\usepackage[preprint,nonatbib]{corl_2026}
\usepackage[T1]{fontenc}
\usepackage{amsmath,amssymb,amsthm}
\usepackage[square,numbers,sort&compress]{natbib}
\usepackage{graphicx}
\usepackage{booktabs}
\usepackage{longtable}
\usepackage{array}
\usepackage{placeins}
\usepackage{microtype}
\usepackage{url}

\newtheorem{theorem}{Theorem}
\newcommand{\R}{\mathbb R}
\newcommand{\norm}[1]{\left\lVert#1\right\rVert_2}
\DeclareMathOperator*{\argmin}{arg\,min}

\makeatletter
\renewcommand{\@noticestring}{LEAP Workshop at CoRL 2026.}
\makeatother

\hypersetup{
 pdfauthor={Doyoung Heo, Chiyoung Kim, Seungyeon Bak, Sumin Shim, Rita Singh},
 pdftitle={Making Task Abstractions Executable: Control-Aware Layout Repair for a Fixed Controller},
 pdfsubject={Task abstractions and control-certified layout repair},
 pdfkeywords={task abstractions, layout repair, closed-loop certification}
}

\title{Making Task Abstractions Executable: Control-Aware Layout Repair for a Fixed Controller}

\author{
  {\normalfont\mdseries
  Chiyoung Kim$^{1,*}$ \quad
  Seungyeon Back$^{2,*}$ \quad
  Sumin Shim$^{3,*}$}\\[0.35em]
  {\normalfont\mdseries
  Doyoung Heo$^{4,*,\dagger}$ \quad
  Rita Singh$^{3,\dagger}$}\\[0.6em]
  $^1$Chung-Ang University \quad
  $^2$Hanyang University \quad
  $^3$Carnegie Mellon University \quad
  $^4$KAIST\\[0.5em]
  {\small $^*$Equal contribution. \quad $^\dagger$Corresponding author.}
}

\begin{document}
\raggedbottom
\maketitle

\begin{abstract}
A task abstraction can specify the intended events while its spatial layout prevents a fixed agent and controller from completing them.
Starting from a supplied structured task record, we compile whole-task tracking, clearance, and actuation requirements into auditable affine layout constraints.
We repair only declared continuous coordinates, preserving event order, timing, topology, and the controller.
A most-violated-row update admits conditional finite-certification and net-displacement bounds; a same-compiler quadratic projection separates the representation from the optimizer.
On three researcher-authored task abstractions, both backends certify all three layouts and complete all 300 fresh paired rollouts per backend.
A risk-target sweep also exposes fixed event tests that the chosen certificate cannot satisfy through layout edits alone.
\end{abstract}


\section{Introduction}
\label{sec:intro}

A task abstraction fixes what a robot should accomplish without determining whether a proposed scene supports that behavior.
Systems that generate tasks or environments from language descriptions make this gap practically relevant~\citep{wang2024gensim,chung2025wrighthere}.
For example, the event sequence switch $\rightarrow$ door $\rightarrow$ goal specifies an order, but a door that accommodates the body may still be too narrow for closed-loop tracking error.
Likewise, correctly ordered targets may require more acceleration than the fixed actuator permits.
The missing condition is not another event in the task: it is a compatible continuous layout.

We study this interface after a structured task specification has been supplied.
The task record fixes event identities and order, predicate types and tolerances, timing, and topology; the agent and controller are also fixed.
Our compiler maps tracking, clearance, and actuator requirements to affine conditions on editable scene coordinates.
These conditions provide an auditable, sufficient test of whether a layout supports the supplied task under the stated control model.
Repair changes the spatial realization, not the declared task.
The task records in our evaluation are researcher-authored; producing them from language or demonstrations is outside our scope.

Our contributions are a control-derived constraint representation, a simple row-update backend with conditional finite-certification and displacement guarantees, and a comparison with quadratic projection under the same compiler.
Three structured tasks isolate door clearance, checkpoint clearance, and fixed-time travel.
We evaluate 1,200 fresh paired method--rollouts and a risk-target sweep, while reporting earlier calibration-selected stress runs separately.

\section{Related Work}
\label{sec:related}

WrightHere connects user descriptions to explorable environments~\citep{chung2025wrighthere}, GenSim generates robot tasks and demonstrations~\citep{wang2024gensim}, and ReShoot augments recorded demonstrations through generative visual domain randomization~\citep{kim2026reshoot}.
Our method instead takes a structured task and initial layout as inputs and repairs their continuous realization for a fixed controller.
Generate--repair also appears in level design~\citep{zhang2020repair}; our tests additionally include feedback tracking, actuator limits, and ordered events.
Funnel and stochastic-safety methods certify closed-loop behavior~\citep{majumdar2017funnels,santoyo2019barrier}; we use a certificate to constrain editable geometry.
The update is related to Motzkin relaxation~\citep{deloera2017sampling}; the contribution is the control-derived compiler and its task--execution interface, not a new projection algorithm.

\section{Problem Formulation}
\label{sec:problem}

The input is a supplied task record and an initial layout.
The record specifies ordered event types, position and speed gate rules, stage durations, fixed identities and topology, and editable spatial fields.
A gate may move with its event location, but its type and tolerance remain fixed.
Let $q\in\R^d$ collect the $d$ editable physical quantities, and let $S\succ0$ be a fixed diagonal matrix of coordinate scales.
We use dimensionless coordinates $\theta=S^{-1}q$, with $\widehat\theta$ denoting the initial layout.
Pinned coordinates and equalities are removed by parameterization; unsupported structural defects are rejected before repair.
Only declared geometry may change: the task record, agent, and controller remain fixed.

Let $z_i(\theta)\in\R^2$ be the affine endpoint of event $i$, $z_0$ the fixed start, and $T_i>0$ its stage duration.
A fixed reference generator joins these endpoints in order.
At control sample $t$, with sampling period $h>0$, the state is $x_t=[p_t^\top\ v_t^\top]^\top\in\R^4$, where $p_t,v_t\in\R^2$ are position and velocity.
Write $x_t^{\mathrm{ref}}(\theta)$ for the reference state and $e_t=x_t-x_t^{\mathrm{ref}}(\theta)$ for tracking error over the $H$-interval horizon.
For an error weight $P=P^\top\succ0$ and allowed whole-task risk $\delta\in(0,1)$, choose $\rho_\delta>0$ such that
\begin{equation}
 \Pr\!\left(
 \max_{0\le t\le H}e_t^\top Pe_t\le\rho_\delta
 \right)\ge1-\delta.
 \label{eq:tube}
\end{equation}
The probability is over initial error and process disturbance for a fixed task and layout.
The compiler combines this tube with body clearance, intersample motion, event gates, parameter bounds, and feedback-input reserve.
Appendix~\ref{app:theory} derives sufficient conditions for ordered completion under the sampled model.

\section{Method}
\label{sec:method}

\paragraph{Certificate compiler.}
The compiler produces affine layout tests and first checks those independent of $\theta$.
A false constant test returns ``uncertified''; if only true constant tests remain, the initial layout is returned.
Otherwise, let $m\ge1$ be the number of nonconstant rows, written $a_j^\top\theta\le b_j$ with $a_j\in\R^d$ and $b_j\in\R$.
For a centered door, let $w(\theta)$ be its half-width, $r_{\rm b}$ the body-plus-clearance radius, and $r_p,r_v$ the position and velocity supports along the door normal.
Its clearance test is $r_{\rm b}+r_p+hr_v\le w(\theta)$.
For coordinate $\ell\in\{1,2\}$, let $U_\ell$ be the input limit and $\eta_\ell$ the feedback reserve.
Stage $i$ must also satisfy
$\pm4[z_{i,\ell}(\theta)-z_{i-1,\ell}(\theta)]/T_i^2+\eta_\ell\le U_\ell$.
These tests are affine; Appendix~\ref{app:compiler} gives their derivation.

Divide each nonconstant row by $\norm{a_j}$, so henceforth $\norm{a_j}=1$.
Define its slack $s_j(\theta)=b_j-a_j^\top\theta$ and the worst-row margin
\begin{equation}
 \mu(\theta)=\min_{1\le j\le m}s_j(\theta).
 \label{eq:margin}
\end{equation}
A layout is certified when \emph{all constant tests pass and} $\mu(\theta)\ge0$.
Compiler soundness and~\eqref{eq:tube} then give
\begin{equation}
 \Pr(\text{task success}\mid\theta)\ge1-\delta.
 \label{eq:success}
\end{equation}
A negative margin withholds this sufficient certificate; it does not predict certain failure.

\paragraph{Repair.}
After the constant checks, initialize $\theta_0=\widehat\theta$ and choose correction headroom $\kappa>0$.
At iteration $k$, choose the least-slack row, breaking ties by lowest index, and use $[z]_+=\max\{z,0\}$ to update
\begin{equation}
 j_k=\argmin_j s_j(\theta_k),\qquad
 \theta_{k+1}
 =\theta_k-[\kappa-\mu(\theta_k)]_+a_{j_k}.
 \label{eq:update}
\end{equation}
The selected row's slack is raised to $\kappa$ when below it; other slacks may decrease.
We recompute all slacks and return the first iterate with $\mu\ge0$, or ``uncertified'' if the budget expires.
Only the returned layout is materialized as $q=S\theta$.
Intermediate updates require no clipping, policy change, or simulator call; a dense iteration, including row selection, costs $O(md)$.

\paragraph{Same-compiler QP backend.}
Let $C\in\R^{m\times d}$ stack the normalized rows $a_j^\top$, and let $b=(b_1,\ldots,b_m)^\top$.
After the same constant checks, QP-Projection uses the same bounds and full rows to solve
\begin{equation}
 \theta_{\rm QP}
 =\argmin_\theta \tfrac12\norm{\theta-\widehat\theta}^2
 \quad\text{s.t.}\quad C\theta\le b.
 \label{eq:qp}
\end{equation}
Its feasible set is $\mathcal F_0=\{\theta:\mu(\theta)\ge0\}$, without $\kappa$ headroom.
Original-row checks and Karush--Kuhn--Tucker (KKT) diagnostics are recorded separately from solver status.
The compiler is shared; the row update and quadratic projection are alternative repair backends.

\section{Analysis}
\label{sec:analysis}

The analysis concerns one fixed affine problem after all constant tests pass.
For headroom $\kappa$, define
$\mathcal F_\kappa=\{\theta:\mu(\theta)\ge\kappa\}$
and the comparison distance
\begin{equation}
 D=\min_{\theta\in\mathcal F_\kappa}
 \norm{\theta-\widehat\theta},
 \label{eq:distance}
\end{equation}
with $D=+\infty$ if $\mathcal F_\kappa$ is empty.
This distance is an analysis quantity, not an input to repair.
Let $N=\min\{k:\mu(\theta_k)\ge0\}$, with $N=+\infty$ if certification never occurs.
For progress diagnostics, continue~\eqref{eq:update} algebraically after $N$; deployment stops at $N$ or its prescribed budget.

\begin{theorem}
\label{thm:main}
Assume all constant tests pass, $m\ge1$ normalized rows remain fixed, the whole-task certificate is sound, and $D<\infty$.
The untruncated update satisfies, for every integer $n\ge1$,
\begin{equation}
 \max_{0\le k\le n}\mu(\theta_k)
 \ge\kappa-\frac{D}{\sqrt n},
 \qquad
 N\le\left\lceil\frac{D^2}{\kappa^2}\right\rceil.
 \label{eq:progress}
\end{equation}
Moreover, for every integer $k\ge0$,
\begin{equation}
 \norm{\theta_k-\widehat\theta}\le2D,
 \qquad
 \Pr(\text{task success}\mid\theta_N)\ge1-\delta.
 \label{eq:finite}
\end{equation}
An initially certified layout is returned unchanged at $N=0$.
\end{theorem}

The progress bound concerns the best-so-far margin, not monotonic improvement of the current margin.
The displacement bound concerns net change, not cumulative edit length or approximation to the closest zero-margin repair.
Finite certification assumes positive-margin feasibility and sufficient iteration budget.
Appendix~\ref{app:proof} proves the bounds; Appendices~\ref{app:control}--\ref{app:compiler} connect the tests to execution.

\section{Experiments}
\label{sec:experiments}

\paragraph{Task inputs.}
We evaluate three researcher-authored task abstractions (Table~\ref{tab:task-main}).
The structured records and initial stress layouts are supplied explicitly.
Each task has three fixed $2$\,s stages with position and speed gates.
Appendix~\ref{app:parameters} distinguishes the descriptions, event records, and spatial defects.

\begin{table}[htb!]
 \caption{\textbf{Supplied task abstractions.} Event identities/order, gate tolerances, and timing remain fixed; only the listed geometry may move.}
 \label{tab:task-main}
 \centering
 \small
 \setlength{\tabcolsep}{4pt}
 \begin{tabular}{@{}llll@{}}
  \toprule
  Task & Fixed event sequence & Isolates & Editable geometry \\
  \midrule
  P1
  & Switch $\rightarrow$ Door $\rightarrow$ Goal
  & Door clearance
  & Door half-width \\
  P2
  & Checkpoint 1 $\rightarrow$ Checkpoint 2 $\rightarrow$ Goal
  & Checkpoint clearance
  & Event $y$-coordinates \\
  P3
  & Switch 1 $\rightarrow$ Switch 2 $\rightarrow$ Goal
  & Fixed-time travel
  & Event $x$-coordinates \\
  \bottomrule
 \end{tabular}
\end{table}

\paragraph{Setup.}
Raw retains the initial layout.
Geometry-only retains range, topology, body-clearance, and nominal-input tests but omits tracking and feedback reserves.
Ours and QP-Projection share the full compiler.
After replaying the earlier calibration-selected stress runs, we use 100 fresh paired initial-error/noise realizations per task, giving $3\times100\times4=1{,}200$ headless MuJoCo method--rollouts~\citep{mujocodocs}.
Repair occurs once per task/method, not per rollout.
Let $\theta^{\rm ret}$ be the returned layout and $\Delta\theta=\theta^{\rm ret}-\widehat\theta$ its edit.
The primary certificate uses $\delta=0.5$.

\begin{figure}[htb!]
 \centering
 \newcommand{\FigOne}[1]{\includegraphics[scale=1.15,clip,trim=#1]%
  {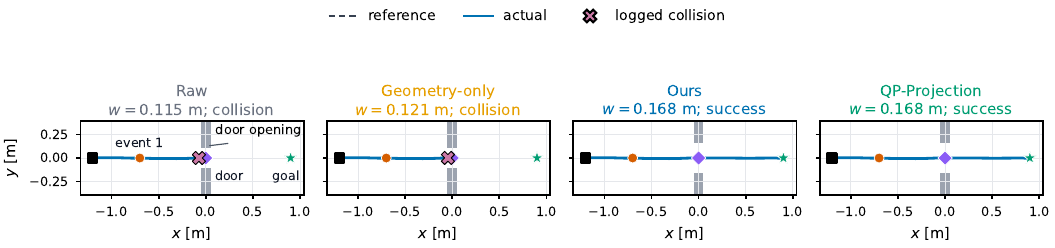}}%
 \FigOne{155 106.2 155.9 2.5}\\[3pt]
 \FigOne{0 2 236.6 39}\\[4pt]
 \FigOne{0 2 471.33 39}
 \FigOne{271.2 2 0 39}
 \caption{\textbf{Matched P1 layouts and logged trajectories.}
 Panels share the task record, controller, and paired disturbance.
 Dashed/solid curves show reference/actual motion; colored points mark ordered events, and the magenta X marks simulator contact.
 Raw and Geometry-only collide; both full-compiler backends widen the door half-width to about $0.168$\,m and succeed.}
 \label{fig:environment}
\end{figure}

\paragraph{Results.}
Both full backends certify all three layouts and complete all 300 fresh rollouts per backend (Table~\ref{tab:results}).
P1/P2 show the execution benefit of tracking-aware clearance.
On P3, Geometry-only succeeds in $99/100$ runs but saturates in every run; full repair instead provides saturation-free execution and certification.
These successful uncertified rollouts show that lacking the certificate does not imply that every execution must fail.
QP makes slightly smaller zero-margin edits.
The measured backend costs are implementation-specific (Appendix~\ref{app:validation}).

\begin{table}[htb!]
 \caption{\textbf{Fresh headless evaluation} ($\delta=0.5$).
 Successes are out of 100 per task.
 Cert.: layouts passing the full compiler; Sat.: rollouts with actuator saturation (all on P3); Edit: mean $\lVert\Delta\theta\rVert_2$.}
 \label{tab:results}
 \centering
 \small
 \setlength{\tabcolsep}{5pt}
 \begin{tabular}{@{}lccccccc@{}}
  \toprule
  & & \multicolumn{3}{c}{Successes (of 100)} & & & \\
  \cmidrule(lr){3-5}
  Method & Cert. & P1 & P2 & P3 & Overall & Sat. & Edit \\
  \midrule
  Raw & 0/3 & 0 & 0 & 0 & 0/300 & 100 & 0 \\
  Geometry-only & 0/3 & 1 & 12 & 99 & 112/300 & 100 & 0.272 \\
  \addlinespace
  Ours & 3/3 & 100 & 100 & 100 & 300/300 & 0 & 0.935 \\
  QP-Projection & 3/3 & 100 & 100 & 100 & 300/300 & 0 & 0.933 \\
  \bottomrule
 \end{tabular}
\end{table}

\paragraph{Certificate limits.}
At $\delta\in\{0.2,0.1,0.05\}$, fixed terminal tests fail before either backend runs (Appendix~\ref{app:delta}).
Layout edits therefore cannot obtain this certificate under the unchanged settings; this is not proof that physical execution is impossible.
The rollouts evaluate three supplied layouts, not the analytical tail bound.

\section{Conclusion}
\label{sec:conclusion}

Control-aware constraints connect supplied task abstractions to sufficient conditions for closed-loop execution. Without changing declared events, timing, or the controller, both the iterative and QP backends repair and certify all three studied layouts and complete all 300 fresh paired rollouts per backend, whereas the Geometry-only baseline completes 112. The guarantee is conditional on the sampled model and a risk target of $\delta=0.5$; at smaller targets, fixed terminal tests fail under the unchanged settings, so layout edits alone cannot obtain the certificate. Future work includes tightening the certificate, allowing structural task changes, and generating task records.
\label{maintextend}

\clearpage
\bibliographystyle{unsrtnat}
\bibliography{main04_references}

\clearpage
\appendix
\numberwithin{equation}{section}
\renewcommand{\topfraction}{0.9}
\renewcommand{\bottomfraction}{0.8}
\renewcommand{\textfraction}{0.08}
\renewcommand{\floatpagefraction}{0.85}

\section{Theory Details and Proofs}
\label{app:theory}

\subsection{Proof of Theorem~\ref{thm:main}}
\label{app:proof}

All statements concern one fixed compiled problem: the same task, agent, controller, risk target, coordinate scales, and normalized rows.
Equality constraints are eliminated before normalization.
A true constant inequality is discarded; a false one stops certification because free coordinates cannot repair it.
If no nonconstant rows remain and all constant checks pass, the proposal is returned immediately.
Below, all constant tests pass and at least one normalized nonconstant row remains.
The proof concerns the untruncated coefficient loop; a smaller practical budget may still return ``uncertified''.

\subsubsection{Distance decrease}

The comparison set $\{\theta:\mu(\theta)\ge\kappa\}$ is a nonempty closed intersection of affine halfspaces.
The squared distance to $\widehat\theta$ is continuous and coercive, so a nearest comparison layout $\theta^\dagger$ exists with distance $D$.
It need not be computed by the repair procedure.

Let $h_k=[\kappa-\mu(\theta_k)]_+$.
If $h_k=0$, the update leaves the candidate unchanged.
Otherwise, the selected row and comparison layout satisfy
\begin{equation}
 b_{j_k}-a_{j_k}^\top\theta_k=\mu(\theta_k),
 \qquad
 b_{j_k}-a_{j_k}^\top\theta^\dagger\ge\kappa.
\end{equation}
Therefore $a_{j_k}^\top(\theta_k-\theta^\dagger)\ge h_k$, and
\begin{align}
 \norm{\theta_{k+1}-\theta^\dagger}^2
 &=\norm{\theta_k-\theta^\dagger-h_ka_{j_k}}^2\\
 &=\norm{\theta_k-\theta^\dagger}^2
 -2h_ka_{j_k}^\top(\theta_k-\theta^\dagger)+h_k^2\\
 &\le\norm{\theta_k-\theta^\dagger}^2-h_k^2.
 \label{eq:app_descent}
\end{align}
For an unnormalized row, a projection scales its residual by the squared row norm.
The unit-normal calculation above therefore relies on the stated normalization.
Equation~\eqref{eq:app_descent} is the standard projection decrease underlying relaxation methods~\citep{deloera2017sampling}, here applied to shifted control-derived scene tests.

\subsubsection{Progress and termination}

Summing~\eqref{eq:app_descent} yields
\begin{equation}
 \sum_{k=0}^{n-1}h_k^2
 \le D^2-\norm{\theta_n-\theta^\dagger}^2
 \le D^2.
\end{equation}
Thus some $k<n$ has $h_k\le D/\sqrt n$.
If its shortfall is zero, its margin is at least $\kappa$; otherwise its margin is at least $\kappa-D/\sqrt n$.
Taking the best candidate proves the margin bound in~\eqref{eq:progress}.

For $D>0$, set $n=\lceil D^2/\kappa^2\rceil$.
Then $D/\sqrt n\le\kappa$, so some candidate is certified and the first such index is no larger than $n$.
If $D=0$, closedness gives $\mu(\widehat\theta)\ge\kappa$ and $N=0$.
An initially certified proposal is also returned at $N=0$ even if its margin is below $\kappa$.
The implemented method may stop between zero and $\kappa$; it does not promise finite-time attainment of $\kappa$.

\subsubsection{Displacement and interpretation}

Equation~\eqref{eq:app_descent} gives $\norm{\theta_k-\theta^\dagger}\le D$.
The triangle inequality then gives
\begin{equation}
 \norm{\theta_k-\widehat\theta}
 \le\norm{\theta_k-\theta^\dagger}
 +\norm{\theta^\dagger-\widehat\theta}
 \le2D.
\end{equation}
This is net layout displacement, not the sum of edit lengths.
It compares with a layout having the stronger margin $\kappa$, not the nearest layout satisfying only $\mu\ge0$.
Consequently, we claim neither globally optimal repair nor an approximation factor relative to the zero-margin problem.
Increasing $\kappa$ changes $D$ and can destroy positive-margin feasibility.
Competing rows can make the current margin nonmonotone; only the best-so-far quantity has the stated guarantee.
Finally, all constant tests pass and $\mu(\theta_N)\ge0$, so compiler soundness and~\eqref{eq:tube} give~\eqref{eq:success} at the returned layout.
This proves the task-success statement without treating margin as a probability.\qed

\subsection{Sampled Control Certificate}
\label{app:control}

This section gives the concrete control model used in the experiments.
It does not assert that arbitrary agents or arbitrary MuJoCo scenes obey this model.

\subsubsection{Sampled dynamics and LQR}

Let $h>0$ be the control period, $p_t,v_t,u_t,w_t\in\R^2$ the planar position, velocity, commanded acceleration, and interval-held acceleration disturbance, and $x_t=[p_t^\top\;v_t^\top]^\top$.
Exact double-integrator sampling gives
\begin{equation}
 x_{t+1}=Ax_t+B(u_t+w_t),\quad
 A=\begin{bmatrix}I_2&hI_2\\0&I_2\end{bmatrix},\quad
 B=\begin{bmatrix}\frac12h^2I_2\\hI_2\end{bmatrix}.
 \label{eq:sampled_model}
\end{equation}
Known mass converts acceleration commands to forces.
References obey
$x_{t+1}^{\mathrm{ref}}=Ax_t^{\mathrm{ref}}+Bu_t^{\mathrm{ref}}$.
For a fixed discrete-time LQR gain $K$, the command
\begin{equation}
 u_t=u_t^{\mathrm{ref}}-Ke_t,\qquad
 e_t=x_t-x_t^{\mathrm{ref}}
\end{equation}
gives ideal error dynamics
\begin{equation}
 e_{t+1}=Fe_t+Bw_t,\qquad F=A-BK.
 \label{eq:error_model}
\end{equation}
Let $Q\succ0$ and $R\succ0$ be the fixed LQR state-error and input-error weight matrices, respectively, of sizes $4\times4$ and $2\times2$.
The stabilizing discrete algebraic Riccati equation (DARE) solution $X$ satisfies
\begin{equation}
 X=A^\top XA
 -A^\top XB(R+B^\top XB)^{-1}B^\top XA+Q.
\end{equation}
The implementation computes $X$ with
\texttt{scipy.linalg.solve\_discrete\_are}
(SciPy 1.17.1, default balancing), forms
$K=(R+B^\top XB)^{-1}B^\top XA$,
and checks that the spectral radius of $F$ is below one.
The solver uses the generalized-eigenvalue/QZ construction of~\citet{vandooren1981riccati}.


Input rows in Appendix~\ref{app:compiler} prevent saturation on the certified event, coupling the physical and ideal processes up to their first possible disagreement.

\subsubsection{Whole-task probability bound}

Let $\mathcal H_t$ denote the history before $w_t$ is applied, so $e_t$ is $\mathcal H_t$-measurable.
Assume $\mathbb E[\norm{e_0}^2]<\infty$ and let $\Sigma\succeq0$ be a finite covariance bound satisfying
\begin{equation}
 \mathbb E[w_t\mid\mathcal H_t]=0,\qquad
 \mathbb E[w_tw_t^\top\mid\mathcal H_t]\preceq\Sigma.
\end{equation}
For $P\succ0$ solving $F^\top PF-P=-I_4$, define
$V(e)=e^\top Pe$ and
$\beta=\operatorname{tr}(PB\Sigma B^\top)$.
Expanding~\eqref{eq:error_model} and using the zero conditional mean gives
\begin{align}
 \mathbb E[V(e_{t+1})\mid\mathcal H_t]
 &=e_t^\top F^\top PF e_t
 +\mathbb E[w_t^\top B^\top PBw_t\mid\mathcal H_t]\\
 &\le V(e_t)-\norm{e_t}^2+\beta
 \le V(e_t)+\beta.
 \label{eq:drift}
\end{align}
For the whole-task horizon $H$, choose
\begin{equation}
 \rho_\delta
 \ge\frac{\mathbb E[V(e_0)]+H\beta}{\delta}.
 \label{eq:rho}
\end{equation}
Define the bounded stopping time
\begin{equation}
 \tau=\min\!\left(
 \{t\in\{0,\ldots,H\}:V(e_t)\ge\rho_\delta\}
 \cup\{H\}
 \right).
\end{equation}
Let $\mathbf 1_{\{\tau>t\}}$ be the indicator that the process has not stopped at time $t$.
Since this indicator is $\mathcal H_t$-measurable, telescoping~\eqref{eq:drift} gives
\begin{align}
 \mathbb E[V(e_\tau)]
 &=\mathbb E[V(e_0)]
 +\sum_{t=0}^{H-1}
 \mathbb E\!\left[
 \mathbf 1_{\{\tau>t\}}
 (V(e_{t+1})-V(e_t))
 \right]\\
 &\le\mathbb E[V(e_0)]
 +\beta\sum_{t=0}^{H-1}\Pr(\tau>t)
 \le\mathbb E[V(e_0)]+H\beta.
\end{align}
On every path reaching the threshold, $V(e_\tau)\ge\rho_\delta$; on all other paths $V(e_\tau)\ge0$.
Therefore
\begin{equation}
 \Pr\!\left(
 \max_{0\le t\le H}V(e_t)\ge\rho_\delta
 \right)
 \le\frac{\mathbb E[V(e_0)]+H\beta}{\rho_\delta}
 \le\delta,
\end{equation}
which implies~\eqref{eq:tube}.
No multiplication of per-stage probabilities or independence between stage successes is used.
The bound is conservative; finite-horizon stochastic safety has broader treatments~\citep{santoyo2019barrier}.

\subsubsection{Exactly realizable references}

Each event has an affine endpoint $z_i(\theta)\in\R^2$; $z_0$ is fixed.
Stage $i$ contains an even number $H_i$ of intervals and duration $T_i=H_ih$.
For local time $s\in[0,T_i]$, use
\begin{equation}
 u_i^{\mathrm{ref}}(s;\theta)=
 \begin{cases}
 \dfrac{4}{T_i^2}
 \bigl(z_i(\theta)-z_{i-1}(\theta)\bigr),
 &0\le s<T_i/2,\\[1mm]
 -\dfrac{4}{T_i^2}
 \bigl(z_i(\theta)-z_{i-1}(\theta)\bigr),
 &T_i/2\le s\le T_i.
 \end{cases}
 \label{eq:two_pulse}
\end{equation}
Integrating from rest reaches $z_i$ at rest exactly after $T_i$.
Position and velocity are continuous at stage transitions, and the actual state is never reset.
An event completes only when its terminal position and speed predicates pass.
The door mode change preserves agent state and applies no impulse; this models an event-controlled door, not physical button manipulation.

\subsection{Constraint Compiler and Numerical Procedure}
\label{app:compiler}

\subsubsection{Space between control samples}

Let $C_p=[I_2\;0]$ and $C_v=[0\;I_2]$ select position and velocity errors.
For wall normal $n$, the ellipsoid support values are
\begin{equation}
 r_p(n)=
 \sqrt{\rho_\delta n^\top C_pP^{-1}C_p^\top n},
 \qquad
 r_v(n)=
 \sqrt{\rho_\delta n^\top C_vP^{-1}C_v^\top n}.
 \label{eq:support}
\end{equation}
For a sample interval starting at $t$, let $e_p(s)$ be the position tracking error after elapsed time $s\in[0,h]$.
The reference and commanded accelerations are held on that interval, so exact integration gives
\begin{equation}
 e_p(s)=C_pe_t
 +s\!\left(1-\frac{s}{2h}\right)C_ve_t
 +\frac{s^2}{2h}C_ve_{t+1}.
\end{equation}
Both velocity coefficients are nonnegative and sum to $s\le h$.
Thus, on the whole-task tracking event,
\begin{equation}
 |n^\top e_p(s)|\le r_p(n)+h\,r_v(n).
 \label{eq:intersample}
\end{equation}
This uses the same event as~\eqref{eq:tube}, so it adds no union bound.
For body-plus-clearance radius $r_{\mathrm b}$ and an affine wall offset $c(\theta)$, the safe wall condition is $n^\top p\le c(\theta)$.
Both endpoints of every segment to which this wall applies must satisfy
\begin{equation}
 n^\top z_i(\theta)
 +r_{\mathrm b}\norm{n}
 +r_p(n)+hr_v(n)
 \le c(\theta).
\end{equation}
The reference stays on the endpoint segment, so these affine endpoint tests contain the swept body.

\subsubsection{Events and actuator reserve}

Target regions use fixed normals and affine offsets with position allowance $r_p(n)$.
For terminal speed limit $v_{\mathrm{gate}}$, the constant test is
\begin{equation}
 \sqrt{
 \rho_\delta\lambda_{\max}(C_vP^{-1}C_v^\top)
 }
 \le v_{\mathrm{gate}}.
\end{equation}
It cannot be repaired by moving a target.
For row $k_\ell^\top$ of $K$ and component limit $U_\ell$, the feedback reserve is
$\eta_\ell=\sqrt{\rho_\delta k_\ell^\top P^{-1}k_\ell}$.
For each stage and both signs, compile
\begin{equation}
 \pm\frac{4}{T_i^2}
 \bigl(z_{i,\ell}(\theta)-z_{i-1,\ell}(\theta)\bigr)
 +\eta_\ell\le U_\ell.
 \label{eq:input}
\end{equation}
These rows prevent clipping on the certified event.
A typed event monitor enables each stage only after the preceding position and speed predicates pass.
When all constant and nonconstant tests pass, every scheduled endpoint satisfies the corresponding event predicates on the whole-task tracking event.
Starting from the first stage, the monitor therefore enables each subsequent stage in order.
Together with the intersample clearance and input bounds, this establishes ordered completion under the stated sampled model and proves~\eqref{eq:success}.
This argument preserves the supplied event specification.

\subsubsection{Normalization, feasible comparisons, and checks}

Parameter ranges and fixed-topology conditions join the row list.
Pinned values and equalities are eliminated before every nonconstant row is divided by its positive coefficient norm.
The implementation uses a conservative $10^{-12}$ offset and independently reevaluates every original inequality with the frozen acceptance tolerance.
The theorem concerns exact inequalities; reported numerical certification uses these stated floating-point checks.
Intermediate candidates are never compiled as scenes.

For theorem diagnostics only, SciPy supplies a feasible comparison point for $\mu\ge\kappa$.
This $\mathcal F_\kappa$ computation is distinct from the zero-margin QP-Projection baseline in~\eqref{eq:qp}.
The comparison point's primal residual is checked at $10^{-8}$, but its available dual-gap diagnostic is too loose to certify an exact optimum in the competing P3 problem.
We therefore denote the verified feasible-point distance by $\bar D\ge D$ and use $\bar D$ in all plotted theorem bounds.
This substitution is conservative: $\lceil\bar D^2/\kappa^2\rceil$ and $2\bar D$ upper-bound the corresponding exact-$D$ quantities.

\subsubsection{MuJoCo agreement}

The MuJoCo family has two slide joints, known mass, direct force input, and enabled contacts.
The controller outputs acceleration and the simulator receives mass times that acceleration.
RK4 uses $0.0025$\,s physics steps, 20 per held $0.05$\,s command.
Across 40 contact-free one-step comparisons, the maximum analytic/simulator state residual was $2.50\times10^{-16}$, below the declared $10^{-9}$ diagnostic tolerance.
This checks the contact-free zero-order-hold implementation, not contact dynamics; collision runs remain empirical.

\section{Experiment Details and Additional Results}
\label{app:experiments}

\subsection{Task descriptions, supplied task records, and fixed parameters}
\label{app:parameters}

Each case has three distinct components: a short task description, a supplied structured event record, and an initial numeric layout.
The descriptions summarize the researcher-authored tasks; the method consumes the structured records rather than parsing the descriptions.
The event record fixes event types and order, position and speed gate rules and tolerances, stage durations, object identities, topology, and editable fields.
An event region follows its editable location; preserving its gate means preserving the predicate form and tolerance, not freezing the region in world coordinates.
The deterministic constructor maps a validated record and layout to MuJoCo's scene format (MJCF), while the repair backends operate only on compiled coefficients.
No language model or automatic semantic parser was used in these experiments.

Table~\ref{tab:prompts} separates the supplied task information from the editable geometry.
Its descriptions are English renderings of the archived researcher-written specifications.
The stress descriptors ``insufficient opening,'' ``too close to walls,'' and ``large travel distance'' describe how the initial layouts were chosen.
They are not task predicates requiring those defects to remain after repair.
Preservation means retaining the declared event record and fixed constraints.

\begin{table}[htb!]
 \caption{Task descriptions and the corresponding supplied task abstractions. The event records are provided inputs.}
 \label{tab:prompts}
 \centering
 \small
 \setlength{\tabcolsep}{6pt}
 \begin{tabular}{
 @{}p{0.04\linewidth}
 >{\raggedright\arraybackslash}p{0.21\linewidth}
 >{\raggedright\arraybackslash}p{0.40\linewidth}
 >{\raggedright\arraybackslash}p{0.24\linewidth}@{}
 }
  \toprule
  ID
  & Task description
  & Fixed task abstraction
  & Editable spatial realization \\
  \midrule
  P1
  & Switch $\rightarrow$ Door $\rightarrow$ Goal.
  & Three ordered monitored events; fixed event endpoints, timing, and event-controlled door logic.
  & Door half-width; the initial opening is insufficient. \\
  \addlinespace
  P2
  & Ordered checkpoints $\rightarrow$ Goal.
  & Checkpoint 1, checkpoint 2, and goal, in that order; fixed horizontal coordinates, corridor, timing, and gate tolerances.
  & Three event heights; the initial targets lie too close to the walls. \\
  \addlinespace
  P3
  & Switch 1 $\rightarrow$ Switch 2 $\rightarrow$ Goal.
  & Three ordered events; fixed start, transverse coordinates, minimum ordering distance, timing, and gate tolerances.
  & Three horizontal event positions; initial travel demands are large relative to time and input limits. \\
  \bottomrule
 \end{tabular}
\end{table}

\paragraph{Event interpretation.}
Each stage ends at a scheduled event location and tests its position and speed predicates.
The monitor permits progression only after the preceding event passes; repair does not remove or reorder events, relax their tolerances, or extend their durations.
The switch and door labels refer to typed events in this planar benchmark.
The door mode change is state-preserving and impulse-free; it does not model physical button manipulation or general contact-rich door-opening skills.

\paragraph{Shared physical and control settings.}
The fixed values are: mass $1.50$\,kg, body radius $0.120$\,m, control period $0.0500$\,s, RK4 physics step $0.00250$\,s, three $2.00$\,s stages, and horizon $H=120$.
The component acceleration limit is $1.40$\,m/s$^2$; terminal position and speed gates are $0.0430$\,m and $0.117$\,m/s.
Initial position/velocity standard deviations are $0.00300$\,m and $0.00500$\,m/s, and interval-held acceleration noise has standard deviation $0.100$\,m/s$^2$.
The primary certificate uses $\delta=0.500$, repair headroom $\kappa=0.00100$, and a 200-update budget.
The complete machine-readable configuration and every compiler row are in the reproducibility archive.


\paragraph{Task-specific coordinates.}
P1 starts at $(-1.20,0)$\,m and has event $x$-coordinates $-0.700$, $0$, and $0.900$\,m.
Its repair changes only the door half-width.
P2 has the same start, fixed $x$-coordinates $-0.400$, $0.400$, and $1.20$\,m, and a $\pm0.450$\,m corridor; only the three event heights may move.
P3 starts at $(0,0)$\,m, uses a $\pm0.550$\,m corridor and $0.300$\,m minimum event ordering, and proposes event positions $1.60$, $3.20$, and $4.80$\,m.
Only those horizontal event positions may move, with the ordered record unchanged.
These are three structured layout instances.

\subsection{Preserved legacy stress evaluation}
\label{app:tasks}

The legacy protocol contains three saved paired realizations per task.
Those conditions were selected during the earlier stress-case calibration and therefore are not an unbiased random sample.
We loaded their archived initial-error and disturbance arrays, reproduced all 27 previous Raw/Geometry-only/Ours records field by field, and evaluated QP-Projection on the same arrays, giving 36 method--rollouts.

\begin{figure}[htb!]
 \centering
 \includegraphics[width=\linewidth]{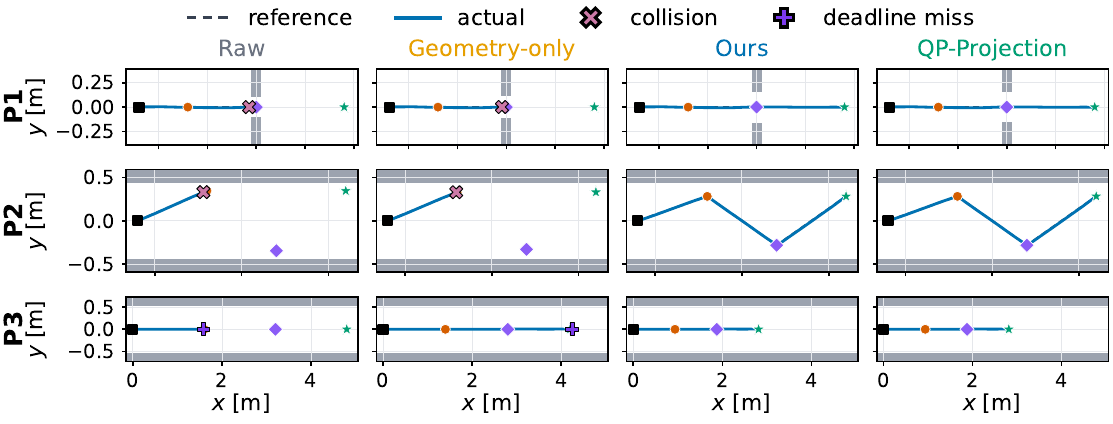}
 \caption{\textbf{All tasks and methods under one preserved paired stress condition.}
 Rows are P1--P3; columns are Raw, Geometry-only, Ours, and QP-Projection.
 Within each row, axes, controller, initial error, and disturbance array are identical.
 Each row has its own $x$ range: the tick labels under the bottom row apply to P3 only, while P1 ticks are at $-1,-0.5,0,0.5,1$\,m and P2 ticks at $-1,0,1$\,m.
 Dashed gray and solid blue curves are reference and actual trajectories; the black square is the start; orange, purple, and green markers are ordered events; magenta X and purple plus mark logged collision and deadline miss.
 Walls and markers are reconstructed from the exact compiled layout, while curves and failure symbols come from simulator logs.}
 \label{fig:all-environments}
\end{figure}

Figure~\ref{fig:all-environments} shows one of these paired conditions for every task and method; columns within a row are directly comparable.
P1 widens the opening, P2 moves three event heights away from the corridor walls, and P3 contracts the three horizontal travel segments.
Raw and Geometry-only terminate at the shown failure symbol in this preserved stress condition; both full-compiler columns reach the green third event.

\begin{table}[htb!]
 \caption{Legacy outcomes after adding QP-Projection. Each task--method cell has three archived paired realizations. Track counts episodes with $\max_t e_t^\top Pe_t/\rho_\delta>1$.}
 \label{tab:legacy}
 \centering
 \small
 \begin{tabular}{@{}llrrrrr@{}}
  \toprule
  Task & Method & Succ. & Coll. & Deadl. & Sat. & Track \\
  \midrule
  P1 & Raw & 0/3 & 3 & 0 & 0 & 3 \\
  P1 & Geometry-only & 0/3 & 3 & 0 & 0 & 2 \\
  P1 & Ours & 3/3 & 0 & 0 & 0 & 0 \\
  P1 & QP-Projection & 3/3 & 0 & 0 & 0 & 0 \\
  P2 & Raw & 0/3 & 3 & 0 & 0 & 3 \\
  P2 & Geometry-only & 1/3 & 2 & 0 & 0 & 0 \\
  P2 & Ours & 3/3 & 0 & 0 & 0 & 0 \\
  P2 & QP-Projection & 3/3 & 0 & 0 & 0 & 0 \\
  P3 & Raw & 0/3 & 0 & 3 & 3 & 3 \\
  P3 & Geometry-only & 2/3 & 0 & 1 & 3 & 1 \\
  P3 & Ours & 3/3 & 0 & 0 & 0 & 0 \\
  P3 & QP-Projection & 3/3 & 0 & 0 & 0 & 0 \\
  \bottomrule
 \end{tabular}
\end{table}

Raw, Geometry-only, and Ours reproduce the earlier $0/9$, $3/9$, and $9/9$ totals exactly; QP-Projection completes $9/9$.
These calibration-selected conditions are retained as stress tests, not relabeled as held-out evaluation.

\subsection{Fresh paired evaluation}
\label{app:fresh}

Before observing outcomes, we froze 100 new realization identifiers per task, disjoint from calibration and legacy identifiers.
For each task/identifier pair, all four methods replay the exact same saved initial error and 120 interval-noise vectors.
Repair is performed once per task/method, so this protocol contains 12 fixed method-layouts and 1,200 method--rollout evaluations---not 1,200 independent task or layout instances.
No identifier was selected or removed after evaluation, and all planned evaluations were executed.

\begin{table}[htb!]
 \caption{Fresh results by task and method. Success confidence intervals are Wilson 95\% intervals in percentage points; Coll., Deadl., Sat., and Track count collision, deadline-miss, saturation, and tracking-threshold episodes.}
 \label{tab:fresh}
 \centering
 \small
 \setlength{\tabcolsep}{5pt}
 \begin{tabular}{@{}llrrrrrr@{}}
  \toprule
  Task & Method & Succ. & 95\% CI
  & Coll. & Deadl. & Sat. & Track \\
  \midrule
  P1 & Raw & 0/100 & [0,3.7] & 100 & 0 & 0 & 89 \\
  P1 & Geometry-only & 1/100 & [0.177,5.45] & 99 & 0 & 0 & 77 \\
  P1 & Ours & 100/100 & [96.3,100] & 0 & 0 & 0 & 0 \\
  P1 & QP-Projection & 100/100 & [96.3,100] & 0 & 0 & 0 & 0 \\
  \addlinespace
  P2 & Raw & 0/100 & [0,3.7] & 100 & 0 & 0 & 86 \\
  P2 & Geometry-only & 12/100 & [7,19.8] & 88 & 0 & 0 & 3 \\
  P2 & Ours & 100/100 & [96.3,100] & 0 & 0 & 0 & 0 \\
  P2 & QP-Projection & 100/100 & [96.3,100] & 0 & 0 & 0 & 0 \\
  \addlinespace
  P3 & Raw & 0/100 & [0,3.7] & 0 & 100 & 100 & 100 \\
  P3 & Geometry-only & 99/100 & [94.6,99.8] & 0 & 1 & 100 & 6 \\
  P3 & Ours & 100/100 & [96.3,100] & 0 & 0 & 0 & 0 \\
  P3 & QP-Projection & 100/100 & [96.3,100] & 0 & 0 & 0 & 0 \\
  \addlinespace
  \bottomrule
 \end{tabular}
\end{table}

In Table~\ref{tab:fresh}, rows within a task share all stochastic arrays but use different fixed references.
Raw fails all tasks.
Geometry-only is nearly always successful on P3 despite saturation in every P3 rollout, but is unreliable on the two clearance tasks and never passes the full compiler.
Ours and QP-Projection complete all 300 rollouts each without any logged failure category.
The 300-trial pooled rates are descriptive statistics for an equal mixture of these three fixed task families; repeated rollouts of one layout do not create independent layout samples or validate the analytical $1-\delta$ probability.

\subsection{Layout edits and QP diagnostics}
\label{app:qp}

Both full backends receive byte-identical normalized and original row arrays, names, categories, bounds, and constant checks for each task.
Geometry-only includes parameter ranges, topology, body clearance, and nominal reference-input feasibility, but excludes tracking and feedback reserves.
Table~\ref{tab:layouts04} audits every returned layout against the full compiler regardless of which backend produced it.

\begin{table}[htb!]
 \caption{All 12 fixed method-layouts. Full margin is the minimum normalized full-compiler slack; original margin independently reevaluates unnormalized full rows. Values are rounded to at most three significant digits; full precision remains in the logs.}
 \label{tab:layouts04}
 \centering
 \small
 \setlength{\tabcolsep}{4.5pt}
 \begin{tabular}{@{}llccccl@{}}
  \toprule
  Task & Method & Full cert.
  & $\lVert\Delta\theta\rVert_2$
  & Full margin & Original margin & Physical edit [m] \\
  \midrule
  P1 & Raw & no & 0 & -0.106 & -0.0528 & 0 \\
  P1 & Geometry-only & no & 0.011 & -0.0947 & -0.0473 & 0.0055 \\
  P1 & Ours & yes & 0.107 & 0.001 & 5.00e-04 & 0.0533 \\
  P1 & QP-Projection & yes & 0.106 & 0 & 5.00e-13 & 0.0528 \\
  \addlinespace
  P2 & Raw & no & 0 & -0.126 & -0.0628 & 0, 0, 0 \\
  P2 & Geometry-only & no & 0.0537 & -0.0947 & -0.0473
  & -0.0155, 0.0155, -0.0155 \\
  P2 & Ours & yes & 0.219 & 0.001 & 5.00e-04
  & -0.0633, 0.0633, -0.0633 \\
  P2 & QP-Projection & yes & 0.218 & 0 & 5.00e-13
  & -0.0628, 0.0628, -0.0628 \\
  \addlinespace
  P3 & Raw & no & 0 & -0.662 & -0.662 & 0, 0, 0 \\
  P3 & Geometry-only & no & 0.751 & -0.461 & -0.461
  & -0.201, -0.401, -0.602 \\
  P3 & Ours & yes & 2.48 & 1.64e-04 & 2.32e-04
  & -0.663, -1.32, -1.99 \\
  P3 & QP-Projection & yes & 2.48 & -9.99e-16 & 9.99e-13
  & -0.662, -1.32, -1.99 \\
  \addlinespace
  \bottomrule
 \end{tabular}
\end{table}

QP-Projection uses SciPy 1.17.1: HiGHS first distinguishes affine infeasibility from numerical failure, then SLSQP minimizes~\eqref{eq:qp} with no backend-specific headroom.
Certification uses the shared $10^{-10}$ acceptance tolerance and an independent original-row/constant audit.
The KKT quantities in Table~\ref{tab:qpdiag} use SLSQP's nonnegative inequality multipliers; they support the numerical projection result but are not a formal exact-optimality proof.

\begin{table}[htb!]
 \caption{QP-Projection diagnostics. Iter. is the SLSQP iteration count; residuals use the constraint matrix $C\theta\le b$. Every solver status was successful, every original-row audit passed, and all displayed KKT diagnostics passed the frozen threshold.}
 \label{tab:qpdiag}
 \centering
 \small
 \begin{tabular}{@{}lrrrrrr@{}}
  \toprule
  Task & Iter. & $\max(C\theta-b)$ & Original res.
  & Stationarity & Complementarity & Signed gap \\
  \midrule
  P1 & 1 & 0 & -5.00e-13 & 2.78e-17 & 0 & 0 \\
  P2 & 2 & 0 & -5.00e-13 & 1.11e-16 & 0 & 6.94e-18 \\
  P3 & 2 & 9.99e-16 & -9.99e-13 & 1.78e-15 & 4.16e-15 & -2.22e-15 \\
  \bottomrule
 \end{tabular}
\end{table}

QP's normalized edit norms are $0.106$, $0.218$, and $2.48$ for P1--P3, slightly below Ours because the iterative backend deliberately corrects a selected row to positive headroom before stopping at nonnegative worst margin.
QP's displacement is not substituted for Theorem~\ref{thm:main}'s distance $D$, which is defined over $\mathcal F_\kappa$ rather than $\mathcal F_0$.

\subsection{Risk-target and certificate sensitivity}
\label{app:delta}

\begin{figure}[htb!]
 \centering
 \includegraphics[width=\linewidth]{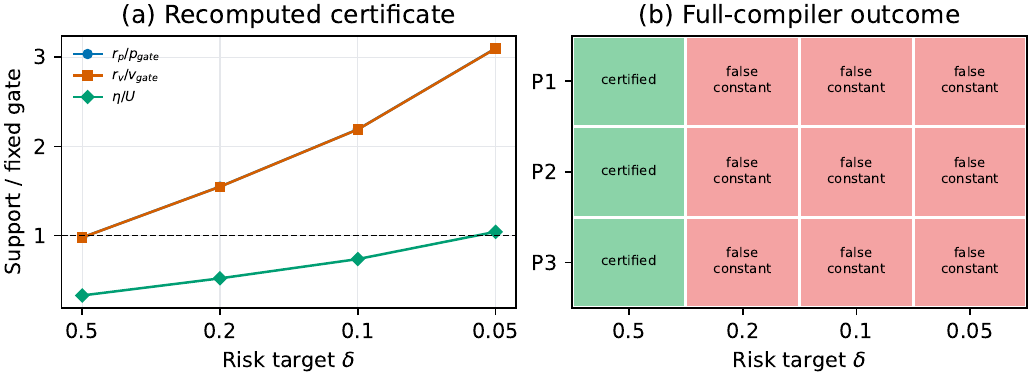}
 \caption{\textbf{Risk-target sweep.}
 Left: recomputed position support $r_p$, speed support $r_v$, and feedback reserve $\eta$, divided by their fixed event/input gates; the dashed line is equality with a gate.
 Right: compiler outcome for each task.
 Green means both constant and affine tests admit certification; red records a false constant test before either repair backend is called.}
 \label{fig:delta}
\end{figure}

We recomputed the complete certificate and compiler for
$\delta\in\{0.500,0.200,0.100,0.0500\}$
while holding dynamics, controller, noise, horizon, task, actuator limits, and event tolerances fixed.
This experiment is coefficient-only: layout repair does not run after a false constant test, and no additional MuJoCo rollout is used.
For the axis-symmetric configuration, write $r_p$, $r_v$, and $\eta$ for the single-axis position support, velocity support, and feedback reserve.
Let $p_{\rm gate}$ and $v_{\rm gate}$ denote the reported terminal position and speed gates, and $U$ the common component input limit used in Figure~\ref{fig:delta}.

In Figure~\ref{fig:delta}, moving right tightens the target risk and enlarges the Markov/Lyapunov tube.
At $\delta=0.500$, both terminal supports fit their fixed gates and all three affine systems are repairable.
At every smaller value, $r_p$ exceeds the $0.0430$\,m position gate and $r_v$ exceeds the $0.117$\,m/s speed gate; these layout-independent failures explain the red cells.
Some tasks additionally fail fixed-direction input tests, but changing event coordinates cannot repair the terminal gates.

\begin{table}[htb!]
 \caption{Certificate values and aggregate compiler outcome. $\eta$ is one axis' feedback-input reserve (the configured controller is axis-symmetric).}
 \label{tab:delta}
 \centering
 \small
 \begin{tabular}{@{}lrrrrl@{}}
  \toprule
  $\delta$ & $\rho_\delta$ & $r_p$ [m] & $r_v$ [m/s]
  & $\eta$ [m/s$^2$] & Outcome (P1--P3) \\
  \midrule
  0.50 & 0.0429 & 0.0421 & 0.115 & 0.462 & certified \\
  0.20 & 0.107 & 0.0666 & 0.181 & 0.73 & false constant test \\
  0.10 & 0.215 & 0.0942 & 0.256 & 1.03 & false constant test \\
  0.05 & 0.429 & 0.133 & 0.362 & 1.46 & false constant test \\
  \bottomrule
 \end{tabular}
\end{table}

Thus the smaller-risk results are classified as \emph{false constant test}, not affine infeasibility, positive-margin infeasibility, iteration-budget exhaustion, or numerical solver failure.
These failures establish that the chosen sufficient certificate cannot be obtained through layout edits alone under the unchanged settings.
They do not establish that the physical tasks are impossible or that their true success probabilities fall below $1-\delta$.
The sweep exposes certificate conservatism without changing event tolerances, timing, or noise to obtain a favorable result.

\subsection{Coefficient-level theorem diagnostics}
\label{app:diagnostics}

For theorem diagnostics, let $\bar\theta\in\mathcal F_\kappa$ be an independently verified feasible comparison layout.
Define its distance from the initial layout by
\begin{equation}
 \bar D=\norm{\bar\theta-\widehat\theta}.
\end{equation}
Then $D\le\bar D$, so replacing $D$ by $\bar D$ in the plotted theorem bounds is conservative; we do not assert that $\bar\theta$ is the exact closest point.
No intermediate repair iterate is simulated.

\begin{figure}[htb!]
 \centering
 \begin{minipage}[t]{0.49\linewidth}
  \includegraphics[width=\linewidth]{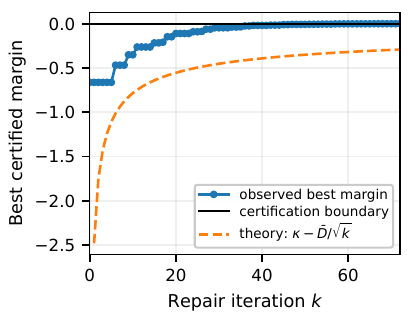}
 \end{minipage}\hfill
 \begin{minipage}[t]{0.49\linewidth}
  \includegraphics[width=\linewidth]{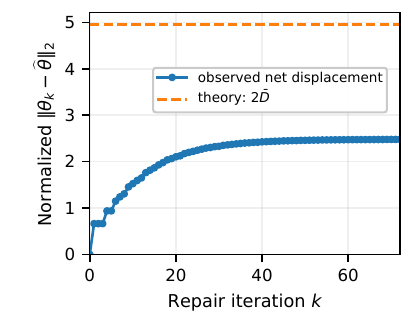}
 \end{minipage}
 \caption{\textbf{P3/Ours coefficient trace.}
 Left: blue markers are logged best-so-far margins; the black line is zero-margin certification and the orange curve is the conservative progress quantity using $\bar D$.
 Right: logged net displacement and the $2\bar D$ upper bound.
 Certification occurs after 72 updates with margin $1.64\times10^{-4}$ and displacement $2.48$.}
 \label{fig:repairdiag}
\end{figure}

Figure~\ref{fig:repairdiag} traces the P3 repair.
Its margin panel concerns the best iterate observed up to each update, not the generally nonmonotone current margin.
Its displacement panel measures straight-line distance from the proposal, not cumulative path length and not a two-approximation to the closest zero-margin repair.

\begin{figure}[htb!]
 \centering
 \includegraphics[width=\linewidth]{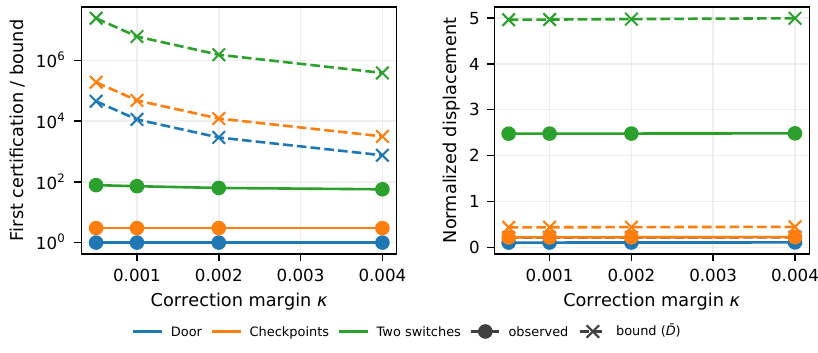}
 \caption{\textbf{Coefficient-only headroom sensitivity.}
 Solid circles are observed first-certification counts or returned displacements for predeclared $\kappa$ multipliers; dashed crosses are conservative bounds recomputed with a feasible comparison for each $\kappa$.
 All 12 repairs certified within budget; no additional simulator episodes were run.}
 \label{fig:kappa04}
\end{figure}

In Figure~\ref{fig:kappa04}, colors identify P1--P3.
The left log-scale panel compares each observed update count with the same-color integer bound; the right panel compares returned displacement with its $2\bar D$ bound.
The gap is conservatism, and the observed decrease in P3 updates as $\kappa$ grows is not claimed as a general monotonicity property.

\begin{figure}[htbp!]
 \centering
 \includegraphics[width=0.76\linewidth]{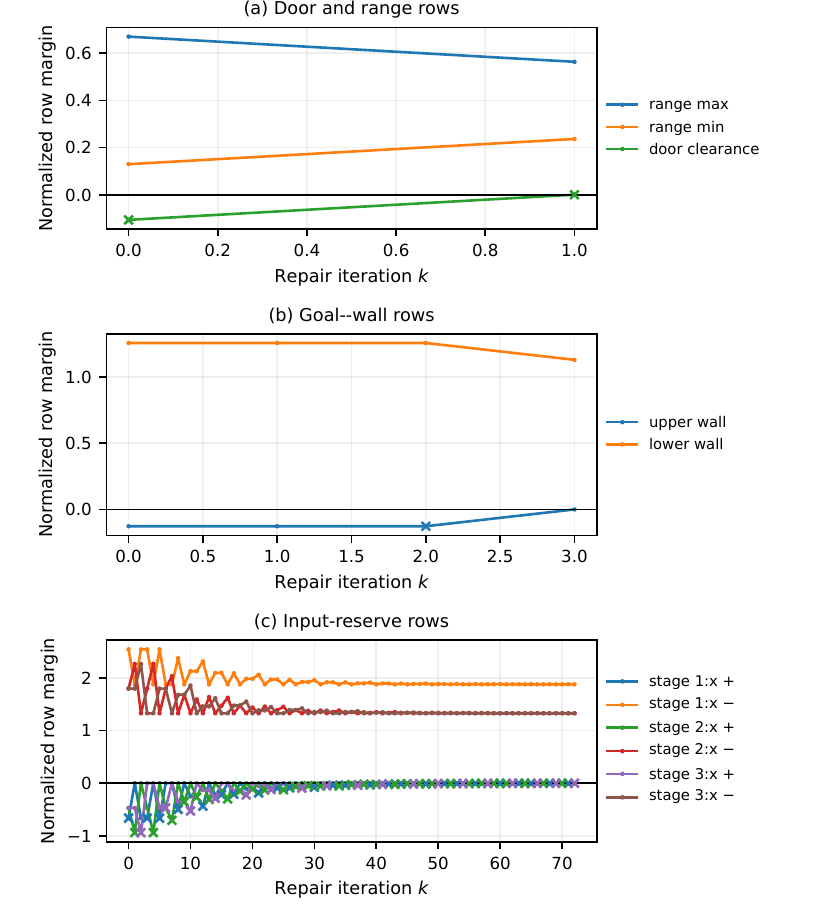}
 \caption{\textbf{Representative row margins for Ours.}
 Curves are named normalized compiler rows; colored crosses mark the row selected as the current worst violation, and the black zero line is certification.
 P1 shows one width correction, while P2/P3 show interactions in which correcting one row can lower another.}
 \label{fig:rowmargins}
\end{figure}

Figure~\ref{fig:rowmargins} follows individual compiler rows: the horizontal axis is the exact coefficient update count and the vertical axis is signed normalized slack.
A curve above zero passes that row, but the scene is certified only when every row and every constant test passes.
Temporary downward motion is an actual constraint interaction, not smoothing.

\subsection{Rollout diagnostics}
\label{app:rollouts}

\begin{figure}[htbp!]
 \centering
 \includegraphics[width=0.85\linewidth]{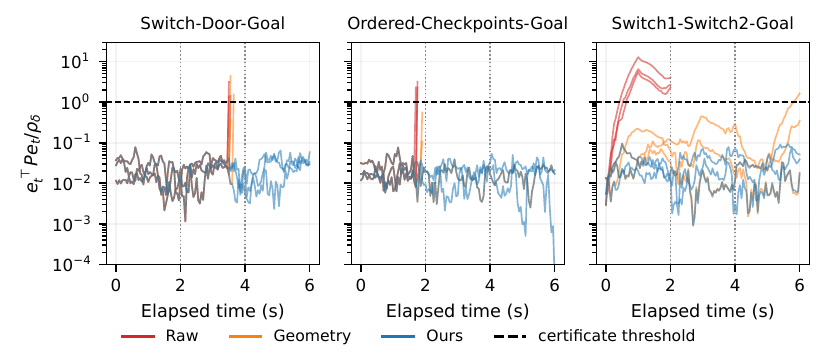}
 \caption{\textbf{Tracking energy in the preserved 27-episode, three-method study.}
 Panels are P1 (Switch-Door-Goal), P2 (Ordered-Checkpoints-Goal), and P3 (Switch1-Switch2-Goal); the legend entry ``Geometry'' is Geometry-only.
 Each faint line is a saved paired realization; red, orange, and blue denote Raw, Geometry-only, and Ours.
 The horizontal dashed line is $e_t^\top Pe_t/\rho_\delta=1$, and vertical dotted lines are the fixed 2 and 4\,s stage boundaries.
 Terminal post-step states are retained, including collision exceedances.}
 \label{fig:tracking04}
\end{figure}

Figure~\ref{fig:tracking04} plots tracking energy for the preserved legacy episodes.
Crossing one is a logged certificate-threshold violation, not automatically a collision.
Early curve termination must be read together with Table~\ref{tab:legacy}: P1/P2 often end at contact, whereas P3 failures are deadline misses following saturation.
The plot predates the QP addition and therefore intentionally shows the exactly reproduced three-method legacy traces; QP's matched outcomes are reported in Table~\ref{tab:legacy} and its full fresh traces remain in the archive.

\subsection{Runtime, reproducibility, and limitations}
\label{app:validation}

At run start, nine identical full episodes benchmarked one, two, and three workers.
All counts were stable.
Although three workers had the highest raw throughput, only $2.59$\,GiB was available, triggering the preset low-memory cap of at most two; one worker was faster than two and was selected for all quantitative runs.
No renderer was constructed during quantitative evaluation and no GPU training or policy training was performed.

\begin{table}[htb!]
 \caption{Measured Windows worker benchmark. Peak RSS is the Python process; selection follows the frozen low-memory policy.}
 \label{tab:workers04}
 \centering
 \small
 \begin{tabular}{@{}rrrrrr@{}}
  \toprule
  Workers & Jobs & Time [s] & Episodes/s & Peak RSS [GiB] & Stable \\
  \midrule
  1 & 9 & 0.255 & 35.3 & 0.115 & True \\
  2 & 9 & 0.284 & 31.6 & 0.117 & True \\
  3 & 9 & 0.225 & 40 & 0.118 & True \\
  \bottomrule
 \end{tabular}
\end{table}

\paragraph{Interpretation of backend timings.}
The recorded row-update calls have lower wall times than the recorded HiGHS--SLSQP--audit pipeline on these three small problems.
These measurements concern the reported implementations and their measured operations, not a solver-independent complexity or speed advantage.
In particular, the QP timing includes a preliminary feasibility solve, constrained minimization, and the original-row audit; it is not the time for the quadratic objective alone.
The benchmark therefore compares the stated backend pipelines rather than matched primitive operations.

\begin{figure}[htb!]
 \centering
 \includegraphics[width=\linewidth]{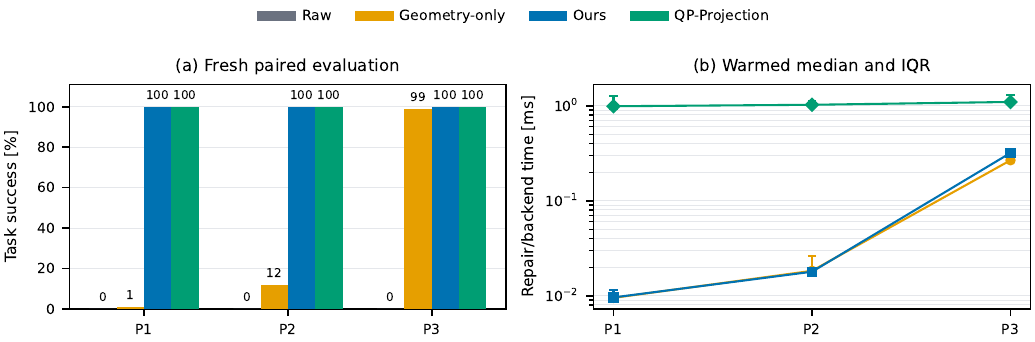}
 \caption{\textbf{Fresh outcomes and recorded backend costs.}
 Left: success over 100 paired realizations per task.
 Right: recorded backend medians and interquartile ranges after 20 warm-ups and over 200 measurements.
 QP includes HiGHS feasibility, SLSQP, and audit; neither timing includes compilation or simulation.
 The operations differ, so this is an implementation-specific timing comparison.}
 \label{fig:fresh}
\end{figure}

\begin{table}[htb!]
 \caption{Separated compiler and backend timings after 20 warm-ups, with 200 measured calls per task/operation. QP includes HiGHS feasibility, SLSQP, and the original-row audit; simulator time is excluded.}
 \label{tab:runtime04}
 \centering
 \small
 \begin{tabular}{@{}llrrr@{}}
  \toprule
  Task & Operation & Median [ms] & IQR [ms] & Repeats \\
  \midrule
  P1 & Full compiler & 0.0393 & 0.00395 & 200 \\
  P1 & Geometry-only & 0.0095 & 5.00e-04 & 200 \\
  P1 & Ours & 0.0096 & 0.00193 & 200 \\
  P1 & QP + audit & 0.987 & 0.322 & 200 \\
  P2 & Full compiler & 0.0715 & 0.00377 & 200 \\
  P2 & Geometry-only & 0.0183 & 0.00835 & 200 \\
  P2 & Ours & 0.0179 & 4.00e-04 & 200 \\
  P2 & QP + audit & 1.02 & 0.118 & 200 \\
  P3 & Full compiler & 0.0611 & 0.0031 & 200 \\
  P3 & Geometry-only & 0.267 & 0.0159 & 200 \\
  P3 & Ours & 0.319 & 0.011 & 200 \\
  P3 & QP + audit & 1.1 & 0.269 & 200 \\
  \bottomrule
 \end{tabular}
\end{table}

The authoritative evaluation run is
\texttt{main04-final-20260922-r3};
its config and evaluation-code SHA-256 prefixes are
\texttt{b6cf58ec2b8c} and \texttt{1447937852b6}.
The fresh 1,200 rollouts took $21.0$\,s and the 36 legacy-plus-QP rollouts took $0.602$\,s.
The machine used Windows 11, an Intel Core Ultra 5 226V (8 cores/8 threads), 16\,GB RAM, Python 3.11.9, NumPy 2.4.6, SciPy 1.17.1, and MuJoCo 3.13.0; BLAS thread environment values were all one.
Eighteen legacy tests and ten QP tests ran inside the final evaluator.

The scope is deliberately narrow: three low-dimensional fixed families, affine editable coordinates, known sampled dynamics, one controller, and a conservative sufficient certificate.
The whole-task probability is analytical under that sampled model; contact-rich MuJoCo completion is empirical and does not validate the tail bound.
A negative margin withholds a sufficient certificate but does not predict certain failure, as Geometry-only's successes show.
The input task abstractions are supplied by researchers.
Task preservation is limited to declared event identities and order, predicate forms and tolerances, timing, and other fixed constraints.
We do not certify arbitrary topology, learned dynamics, physical button manipulation, or online replanning.

\end{document}